\documentclass[letterpaper, 10 pt, conference]{ieeeconf}  

\IEEEoverridecommandlockouts                              

\usepackage{times}
\usepackage{epsfig}
\usepackage{graphicx}
\usepackage{amsmath}
\usepackage{amsfonts} 
\usepackage{hyperref}
\usepackage{cite}

\usepackage{amssymb}
\usepackage{algorithm}
\usepackage[noend]{algpseudocode}
\usepackage{csvsimple}
\usepackage{enumerate}

\usepackage{caption}
\usepackage{subcaption}
\DeclareCaptionSubType*[Alph]{table}
\DeclareCaptionLabelFormat{mystyle}{Table~\bothIfFirst{#1}{ ̃}#2}
\usepackage{multirow}
\usepackage{array}
\newcolumntype{P}[1]{>{\centering\arraybackslash}p{#1}}

\usepackage{mathtools}

\usepackage{enumerate}

\newcommand{\Acondition}{\emph{CBCL} condition}
\newcommand{\Bcondition}{\emph{FT} condition}

\usepackage{xcolor}
\usepackage{comment}

\title{\LARGE \bf
How Do Human Users Teach a Continual Learning Robot\\ in Repeated Interactions?
}

\author{Ali Ayub$^{1*}$, Jainish Mehta$^{1}$, Zachary De Francesco$^{1}$, Patrick Holthaus$^{2}$,
Kerstin~Dautenhahn$^{1}$ \\and Chrystopher L.\ Nehaniv$^{1}$
\thanks{
This research was undertaken, in part, thanks to funding from the Natural Sciences and Engineering Research Council of Canada (NSERC) and the Canada 150 Research Chairs Program.}
\thanks{$^{1}$University of Waterloo, Waterloo, Ontario N2L 3G1, Canada}
\thanks{{\tt\small $\{$*\,a9ayub, jm3mehta, zdefrancesco, cnehaniv, kdautenh$\}$
@uwaterloo.ca}}
\thanks{$^{2}$University of Hertfordshire, Hertfordshire AL10 9AB, England, UK}
\thanks{{\tt\small p.holthaus@herts.ac.uk}}
}

\begin{document}

\maketitle
\thispagestyle{empty}
\pagestyle{empty}







\begin{abstract}
\label{sec:Abstract}
Continual learning (CL) has emerged as an important avenue of research in recent years, at the intersection of Machine Learning (ML) and Human-Robot Interaction (HRI), to allow robots to continually learn in their environments over long-term interactions with humans. Most research in continual learning, however, has been \textit{robot-centered} to develop continual learning algorithms that can quickly learn new information on static datasets. In this paper, we take a \textit{human-centered} approach to continual learning, to understand how humans teach continual learning robots over the long term and if there are variations in their teaching styles. We conducted an in-person study with 40 participants that interacted with a continual learning robot in 200 sessions. In this between-participant study, we used two different CL models deployed on a Fetch mobile manipulator robot. An extensive qualitative and quantitative analysis of the data collected in the study shows that there is significant variation among the teaching styles of individual users indicating the need for personalized adaptation to their distinct teaching styles. The results also show that although there is a difference in the teaching styles between expert and non-expert users, the style does not have an effect on the performance of the continual learning robot. Finally, our analysis shows that the constrained experimental setups that have been widely used to test most continual learning techniques are not adequate, as real users interact with and teach continual learning robots in a variety of ways. Our code is available at \url{https://github.com/aliayub7/cl_hri}.
\end{abstract}

\section{Introduction}
\label{sec:introduction}
\noindent
We envision a future of general-purpose assistive robots that can help users with a variety of tasks in dynamic environments, such as homes, offices, etc. It would be necessary that such assistive robots are personalized to their users' needs and their environments~\cite{Saunders16}. However, over the long term, users' needs, preferences, and environments will continue to change, which makes it impossible to pre-program the robot with all the tasks it might be required to perform. A solution to this problem is to allow people to continually teach their robots new tasks and changes in their environments on the fly, an approach known as continual learning (CL) \cite{Dehghan19,Ayub_IROS_20}. 

Continual learning has been extensively studied in recent years to allow robots to learn over long periods of time \cite{Rebuffi_2017_CVPR,Ayub_IROS_20}. As it is imperative for a robot to learn the objects in its environment, the majority of research on CL has focused on machine learning (ML) models for object recognition in recent years \cite{hayes20,Rebuffi_2017_CVPR,ayub2022fewshot}. Most of these techniques were tested on static object recognition datasets with a large number of training images for each object class. In real-world environments, however, robots will need to learn from individual interactions with their users who might be unwilling to provide a large number of training examples for each object. 

In the past few years, robotics researchers developed CL techniques that can learn from only a few training examples per object, an approach known as Few-Shot Class Incremental Learning (FSCIL) \cite{Ayub_IROS_20,Tao_2020_CVPR,Zhang_2021_CVPR}. Although FSCIL techniques produced promising results on real robots, they were only tested with systematically collected datasets by their experimenters. Overall, most research in continual learning has been \textit{robot-centered}, to develop efficient CL algorithms that can learn from static datasets or interaction with robot experimenters. However, in the real world, robots will learn from real users who might be unfamiliar with robot programming and learning. Therefore, an equally important area of research in continual learning is \textit{human-centered}, to understand how human users interact with and teach continual learning robots over the long term. To the best of our knowledge, we know of no other work on developing long-term user studies where human users teach modern CL models deployed on robots over multiple interactions. 

In this paper, we have a human-centered focus to uncover the diversity and evolution of human teaching when interacting with a continual learning robot over repeated sessions. We developed a CL system that integrates a graphical user interface (GUI) with CL models of object learning deployed on the Fetch mobile manipulator robot \cite{Wise16}. We conducted a long-term between-participant study (N=40) where participants interacted with and taught everyday household objects to a Fetch robot that used two different CL models. We analyzed the data collected in the study to characterize various aspects of human teaching of a continual learning robot in an unconstrained manner. Our results highlight the variation in the teaching styles of different users, as well as the influence of the robot's performance on users' teaching styles over multiple sessions. Our results indicate that the constrained experimental setups traditionally used to test most CL models are inadequate, as real users teach continual learning robots in a variety of ways. 

\section{Related Work}
\label{sec:related_work}
\noindent
In this section, we first present an overview of modern CL methods mostly tested without human users, and then introduce current approaches to robot teaching, highlighting the need for a human-centered approach at the intersection of CL and human-robot interaction (HRI).
\subsection{Continual Learning}
\noindent
The goal of CL models is to continuously adapt and learn new information over time while preserving past knowledge. Most research in the CL literature has focused on class-incremental learning (CIL) in which a machine learning model learns from labeled training data of different classes in each increment and is then tested on all the classes it has learned so far \cite{Rebuffi_2017_CVPR}. One of the main problems faced by class-incremental learning models is \textit{catastrophic forgetting}, in which the model completely forgets the previously learned classes when learning new classes in an increment \cite{french19}. Various research directions have been pursued in the past to tackle the catastrophic forgetting problem, such as replay-based techniques that store and replay data of the old classes when learning new classes \cite{Rebuffi_2017_CVPR,Wu_2019_CVPR}, regularization techniques \cite{kirkpatrick17,Li18}, and generative replay based techniques that generate old data using stored class statistics \cite{ayub2021eec,Ostapenko_2019_CVPR}. These techniques, however, are not suitable for learning from human users who might be unwilling to provide hundreds or thousands of images per object class. 

In the past couple of years, researchers also developed class-incremental learning models that can learn from only a few labeled examples per class, a direction known as few-shot class incremental learning (FSCIL) \cite{tao_2020_ECCV}. However, CIL and FSCIL approaches were either tested on static datasets, or on data captured by a robot while interacting with experimenters in systematically controlled setups \cite{tao_2020_ECCV,Ayub_IROS_20,Dehghan19}. To the best of our knowledge, all of the FSCIL approaches were robot-centered and none of these approaches were tested with actual participants (users).

\begin{figure}[t]
\centering
\includegraphics[width=\linewidth]{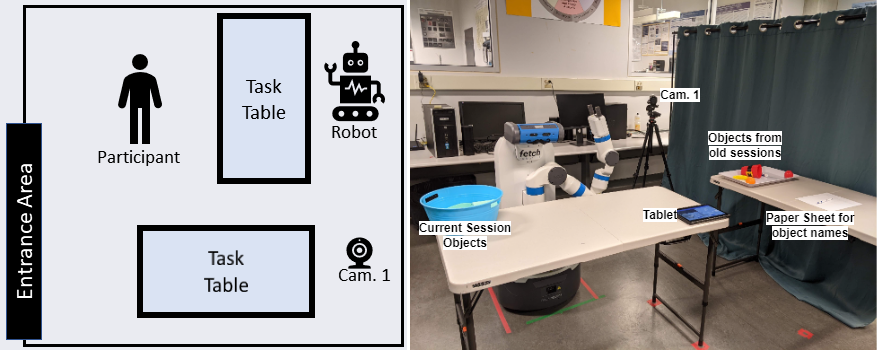}
\caption{\small (Left) Experimental layout for the CL setup with the participant and the robot. (Right) Corresponding real-world setup.}
\label{fig:experimental_setup}
\end{figure}

\subsection{Human-Robot Teaching}
\noindent Human-centered research for robot learning through HRI has been limited. A few user studies have been conducted in the past with simulated and real robots to understand the characteristics of human teaching. Most of these studies were conducted in Wizard of Oz setups where the robot did not learn from human teaching \cite{ramaraj2021unpacking,kaochar2011towards}. Some research has been conducted on interactive reinforcement learning through HRI for learning manipulation tasks through physical human corrections, learning kitchen-related tasks in simulation, or learning natural language description of images from humans \cite{bajcsy2018learning,senft2017supervised,krishna2022socially}. However, most of these studies were designed to test the performance of the reinforcement learning models or understand the perceptions of users towards these models and were not focused on understanding patterns of human teaching. Furthermore, these studies were only tested in a single interaction with users. However, for continual learning robots, it is imperative to design multi-session studies to understand how human teaching of continual learning robots evolves over the long term. 
In contrast to prior work, to the best of our knowledge, we conducted the first long-term user study at the intersection of continual machine learning and HRI, to understand patterns of human teaching with a continual learning robot over multiple interactions.

\section{Method}  
\noindent
We investigated human teaching patterns when interacting with a continual learning robot to teach an object recognition task. The subsections below describe our CL system and the method for our long-term study. 

\subsection{Continual Learning System}
\label{sec:cl_system}
\noindent
In this experiment, in each session, the user taught the robot household objects in a table-top environment and then tested the robot to find and point to the requested object on the table. Figure \ref{fig:experimental_setup} shows the table-top experimental setup for this study. The simplicity of the setup and the task makes it clear what the user should do to teach the robot different objects, and what the robot should do to find the learned objects during the testing phase.  

\begin{figure}[t]
\centering
\includegraphics[width=\linewidth]{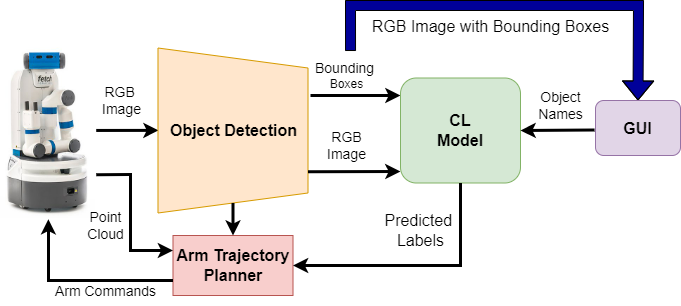}\caption{\small Our complete CL system. Processed RGB images from the robot's camera are sent to the GUI for transparency and also passed on to the CL Model. The user sends object names to the CL model either for training the CL model or finding an object. The arm trajectory planner takes point cloud data, processed RGB data, and predicted object labels from the CL model as input and sends the arm trajectory for the Fetch robot to point to the object.}
\label{fig:sgcl_architecture}
\end{figure}

\begin{figure}[t]
\centering
\includegraphics[width=0.9\linewidth]{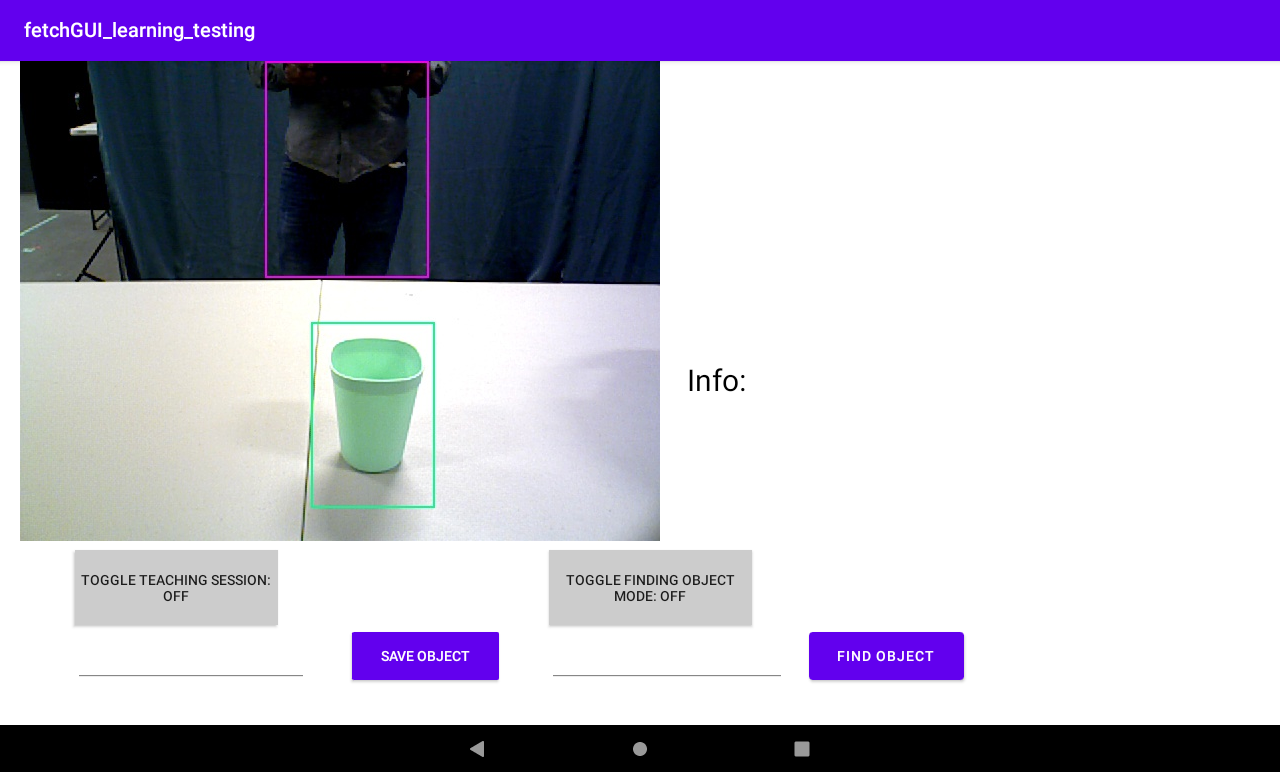}
\caption{\small The graphical user interface (GUI) used to interact with the robot. The RGB camera output with bounding boxes is on the top left. The buttons at the bottom can be used to teach objects to the robot and ask it to find objects in the testing phase. The top right of the GUI shows information sent by the robot to the user.}
\label{fig:gui}
\end{figure}

For this setup, we developed a CL system for the object recognition task, which integrates CL models with a Fetch mobile manipulator robot \cite{Wise16}, as well as a graphical user interface (GUI) for interactive and transparent learning from human users. Figure \ref{fig:sgcl_architecture} shows our system for the object recognition task. In this system, the user interacts with the robot through the GUI on an Android tablet (Figure \ref{fig:gui}). The user provides labels of new objects placed in front of the robot through the GUI and saves the images of objects processed through the object detection module in the robot's memory. The robot then uses the saved object images in each session to train the CL model. After teaching, the user can test the robot by asking it to find objects on the table through the GUI. The robot passes the pre-processed images to the CL model to get the predicted object labels. If the object requested by the user is found, the robot finds the 3D location of the object on the table and points to the object using its arm.

\subsubsection{Continual Learning Models}
\label{sec:cl_models}
We consider two CL models in this study. For the first model, we consider a na{\"\i}ve finetuning (FT) approach \cite{Rebuffi_2017_CVPR} in which a convolutional neural network (CNN) \cite{He_2016_CVPR} is trained on the image data of the object classes in each increment (i.e.\ in an interactive session with the user). The model does not train on any of the objects learned in the previous increments (sessions) and therefore it forgets the previously learned objects. This model can serve as a baseline for forgetting in continual learning \cite{Rebuffi_2017_CVPR,Wu_2019_CVPR}.

For the second model, we consider a state-of-the-art CL approach specifically designed for FSCIL in robotics applications \cite{Ayub_IROS_20}. This approach, termed centroid-based concept learning (CBCL), mitigates forgetting by creating separate clusters for different object classes. CBCL stores cluster centroids of object classes in memory and uses these centroids to make predictions about labels of new objects. More details about these models can be found in \cite{Rebuffi_2017_CVPR,Ayub_IROS_20}. Note that all of these models were only tested on systematically collected object datasets in prior work, and have never been tested in real-time with human participants. 

\subsection{Participants}
\label{sec:participants}
\noindent
We recruited 40 participants (19 female (F); 21 male (M), all students) from the University of Waterloo, between the ages of 18 and 37 years ($M=23.48$, $SD=4.49$). 20 participants (ages: $M=24.15$, $SD=4.21$, 10 F, 10 M) were randomly assigned to \Bcondition{}, and the other 20 (ages: $M=22.78$, $SD=4.68$, 9 F, 11 M) were randomly assigned to \Acondition{}.  
The participants had diverse backgrounds in terms of their majors, but most of them (65\%) were engineering and computer science students. Based on their self-assessments in a pre-experiment survey, 40\% of the participants reported that they were familiar with robot programming, 55\% reported that they had previously interacted with a robot, 5\% were familiar with the Fetch robot, and 10\% had previously participated in an HRI study. For the remainder of the paper, we will call 40\% participants that had prior robot programming experience `experts' and the rest of the participants `non-experts'. 
All procedures were approved by the University of Waterloo Human Research Ethics Board. 

\subsection{Research Questions}
\label{sec:research_questions}
\noindent We analyze the data collected in our study to answer the following research questions and test the associated hypotheses. These hypotheses are guided by previous research that was  discussed in Section \ref{sec:related_work}:  Prior HRI research showed that users' interactions and perceptions towards a robot are correlated with the performance of the robot and the time and effort spent in interacting with the robot. Also, there might be differences in how different users interact with the robot, especially if they had prior experience programming robots. Further, prior CL research showed that CL models can forget previous knowledge over time, and thus their performance decreases. However, there is a difference in the rate of forgetting for different CL models. 

\begin{itemize}
    \setlength{\itemindent}{0.65em}
    \item [\textbf{RQ1}] How do different human users label objects when teaching a continual learning robot over multiple sessions?

        \item [\textbf{H1.1}] Labelling strategies for objects vary among different users.\\[-.5em]
    \item [\textbf{RQ2}] Does the continual learning robot's performance affect the way users teach over multiple sessions?
        \item [\textbf{H2.1}] Classification performance of the robot affects the teaching style of the participants over multiple sessions.
        \item [\textbf{H2.2}] Users teach a robot that forgets previous objects differently than a robot that remembers previous objects.\\[-.5em]
    \item [\textbf{RQ3}] Do users change the way they teach the continual learning robot over multiple sessions?
        \item [\textbf{H3.1}] Teaching styles of users change over multiple sessions regardless of the CL model.\\[-.5em]
    \item [\textbf{RQ4}] Is there a difference in teaching style and robot performance for expert and non-expert users?
        \item [\textbf{H4.1}] Continual learning robots taught by expert users perform better than the ones taught by non-expert users.
        \item [\textbf{H4.2}] There is a difference between the teaching styles of expert and non-expert users.
\end{itemize}

\subsection{Procedure}
\noindent
We conducted five repeat sessions (each lasting $\sim$20-30 minutes) with each participant in a robotics laboratory. All sessions were video recorded. We also stored the image data of the objects taught and tested by the participants. Each participant was randomly assigned to one of the two experimental conditions using one of the two CL models, CBCL and FT. Before their first session, each participant was asked to complete a consent form and a pre-experiment survey online. After completing the consent form and the pre-experiment survey, the experimenter greeted the participant and gave a brief oral introduction to the experiment. The participant then interacted with the robot in a demo session to understand how to teach and test the robot. In the demo phase, the robot did not learn any objects. 

During the demo phase, the experimenter explained to the participant how to start a teaching session using the GUI, teach an object to the robot, and test the robot to find the object. The participant then tried teaching a demo object (this object was not used later) to the robot. The participant then tested the robot to find the demo object on the table using the GUI. After the demo phase ($\sim$5 minutes), the experimenter gave a paper sheet, which served as a memory aid, to the participant to write down the names of the objects of the current session. In this way, the participants could remember the object names when they needed the robot to find these objects in the next sessions. The experimenter then took the tablet from the participant and loaded the program for the actual session on the tablet. The experimenter handed the tablet back to the participant and placed five objects to be taught in the session on one side of the table. The experimenter then mentioned to the participant that they can start their session and start teaching the five objects.

\begin{figure}[t]
\centering
\includegraphics[width=0.45\linewidth]{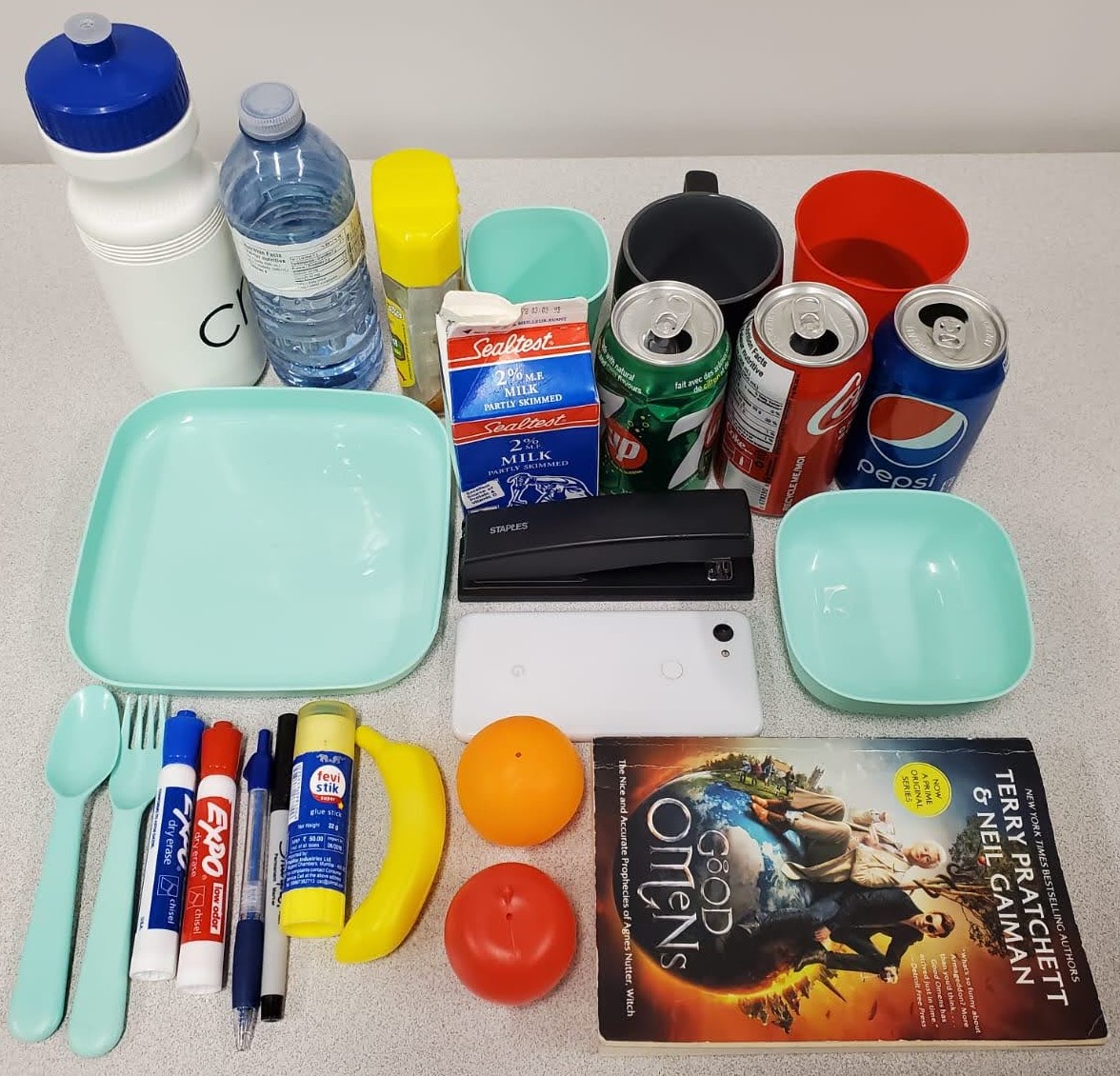}
\caption{\small The twenty-five objects used in our study.} 
\label{fig:objects}
\end{figure}

The experimenter then went to a secluded area and the participant taught and tested objects to the robot. At the end of the session, the experimenter came out of the secluded area and asked the participant to finish a post-experiment survey. The participant then scheduled their next session. In the next four sessions, the same procedure was repeated, except for replacing the objects to be taught between sessions. Figure~\ref{fig:objects} shows the 25 objects used in our study.
Participants were also told that they can bring a maximum of two objects per session of their own choice in sessions 3-5 to teach to the robot. If participants brought their own objects, we replaced some of the objects from our set (Figure~\ref{fig:objects}) with participants' objects (total objects taught over 5 sessions was still 25). Participants did not go through a demo interaction in the next four sessions. At the end of the last session, the experimenter asked the participant to have a short interview to answer some questions describing their experience with the robot. This interview was audio recorded. Analyses of the post-experiment survey and audio interview are not reported since they go beyond the scope of this paper, and will be reported in future publications. Examples of the teaching and testing phases are shown in the supplementary video.

\subsection{Measures}
\noindent We used both qualitative and quantitative measures to analyze the data for the two conditions. 
We analyzed the object names given by the participants to different objects using the image data stored for objects during teaching sessions. We report the variety and frequency of labels used by the participants for each object. We also coded the video recordings to calculate the frequency of teaching by the participants in all 5 sessions, and if they re-taught any objects to the robot in case the robot was not able to correctly find them on the table. 

We also analyzed the performance of two CL approaches. Classification accuracy per session (increment) has been commonly used in the CL literature \cite{Rebuffi_2017_CVPR,Tao_2020_CVPR} for quantifying the performance of CL models for object recognition tasks. Therefore, for each session, during the testing phase, we recorded the total number of objects tested by the participant and the total number of objects that were correctly found by the robot. Using this data, we calculated the accuracy $\mathcal{A}$ of the robot in each session as: \begin{equation}
\mathcal{A} = \frac{number\ of\ objects\ correctly\ found}{number\ of\ objects\ tested}
\end{equation}

\begin{table}[t]
\centering
\begin{tabular}{P{1.5cm}|P{2.9cm}|P{2.7cm}}
    \textbf{Object} & \textbf{No. of Different Labels} & \textbf{Most Common Label} 
    \\
     \hline
     Green Cup & 10 & Cup (59\%)
     \\
     Honey & 13 & Honey (46.5\%)
     \\ 
     Bowl & 10 & Bowl (65\%)
     \\
     Glue & 6 & Glue (76\%)
     \\
     Spoon & 6 & Spoon (81\%)
     \\
     \hline
     Apple & 3 & Apple (90\%)
     \\
     Banana & 3 & Banana (90\%)
     \\
     Red Cup & 14 & Red Cup (25\%)
     \\
     Blue Marker & 11 & Marker (58\%)
     \\
     Orange & 5 & Orange (77\%) 
     \\
     \hline
     Mug & 7 & Mug (72\%) 
     \\
     Fork & 6 & Fork (76\%) 
     \\
     Sharpie & 8 & Sharpie (48\%) 
     \\
     Plate & 10 & Plate (61\%) 
     \\
     Stapler & 6 & Stapler (86\%) 
     \\
     \hline
     Book & 4 & Book (86\%) 
     \\
     Red Marker & 4 & Red Marker (31\%) 
     \\
     Blue Pen & 7 & Pen (60\%) 
     \\
     Pepsi & 7 & Pepsi (54\%) 
     \\
     White Bottle & 8 & Water Bottle (62\%) 
     \\
     \hline
     Coca Cola & 8 & Coke (36\%) 
     \\
     Milk & 7 & Milk (77\%) 
     \\
     Phone & 5 & Phone (68\%) 
     \\
     7Up & 12 & 7Up (44\%) 
     \\
     Water Bottle & 8 & Water (47\%) 
     \\
 \end{tabular}
  \caption{\small The number of different labels given by the participants to all 25 objects in the study together with the most common label for each object with the percentage of participants that chose this label. Objects are ordered from top to bottom as they were taught in 5 sessions with 5 objects per session. Note that the first column shows some reference names for the objects to be able to identify them individually in the paper.} 
  \label{tab:object_names}
 \end{table}

\begin{table*}
    \centering
    \begin{center}
    \begin{tabular}{P{6.5cm}|P{1.8cm}|P{1.7cm}|P{2.3cm}|P{1.5 cm}}
        ~ & \textbf{Accuracy} & \textbf{No. images} & \textbf{Teaching phases} & \textbf{Reteaching} \\ \hline
        Session number & \textbf{0.0003} (0.08) & 0.056 & 0.775 & 0.407 \\ 
        CL model & \textbf{$<$0.0001} (0.32) & 0.161 & 0.774 & 0.748 \\ 
        Programming experience & 0.487 & \textbf{0.034} (0.09) & 0.486 & 0.945 \\ 
        Session number : CL model & \textbf{$<$0.0001} (0.09) & 0.194 & 0.534 & 0.335 \\ 
        Session number : Programming experience & 0.392 & \textbf{0.039} (0.01) & 0.239 & 0.341 \\ 
        CL model : Programming experience & 0.597 & \textbf{0.040} (0.08) & 0.379 & 0.375 \\ 
        Session number : CL model : Programming experience & 0.964 & 0.091 & 0.739 & 0.136 \\ 
    \end{tabular}
    \end{center}
    \caption{Results ($p$ values) of the three-way ANOVA using session number, continual learning model, and previous programming experience as independent variables. Columns for the accuracy of the models, number of images per object, number of teaching phases, and reteaching misclassified objects show $p$ values for the dependent variables. Effect sizes (generalized eta \cite{lakens2013}) for significant ANOVAs are shown in the brackets. Significance levels ($* := p < .05; ** := p < 0.01; *** :=p < 0.001; ****:=p<0.0001$).}
    \label{tab:anova_results}
\end{table*}

We use the accuracy of the models to determine the teaching quality of the participants in each condition and over multiple sessions. Further, using the image data stored for the objects, we calculated the average number of times each object was taught by the participants in each session to determine the effort spent by the participants in teaching the robot. Finally, we analyze how the above-mentioned variables are affected by the sessions, choice of the CL model, and previous robot programming experience of the participants.

\section{Results}
\label{sec:experiments}
\noindent In this section we present the results of our analysis in terms of different labeling strategies and teaching styles of the participants. We also report the effect of participants' teaching styles on the robot's performance and vice versa. 

\subsection{Object Labeling by Human Teachers}
\noindent Table~\ref{tab:object_names} shows the number of different labels given to the 25 objects by 40 participants in the study. To identify each object we add a generic name for each object in the table. For example, for the plastic apple used in our study, we identify it as an apple in the table. Overall, there was a significant variation in the labeling of objects by the participants, ranging from 3 (for Apple) to 14 (for Red Cup) different labels for the objects. Among such labels, some were quite simple and generic, such as \textit{Honey}, \textit{Bowl}, \textit{Milk}, etc. whereas some were quite specific, such as \textit{Almost Empty Yellow Honey Jar}, \textit{Light Green Flat Bowl}, \textit{Empty Milk Carton}, etc. We also report the most common label given to each object and the percentage of participants that chose that label. The consensus among the participants for labeling the objects varied from 25\% for \textit{Red Cup} to 90\% for \textit{Apple}. 

We also noticed some unique labeling strategies by the participants. Some participants labeled different objects in different sessions using the same label. For example, multiple participants gave the label \textit{Cup} to \textit{Green Cup} in Session 1, \textit{Red Cup} in Session 2, and \textit{Mug} in Session 3. In total, 10 out of 40 participants (25\%) gave the same label to at least two different objects. Further, some participants also gave multiple labels to the same objects. For example, one participant labeled \textit{Milk} as both \textit{Milk Box} and \textit{Milk Pouch}. Overall, there were 7 out of 40 participants (17.5\%) that gave more than one label to at least one object. 
Finally, we noticed that some participants gave labels that did not match the objects. For example, one participant named glue \textit{Insert Stick Joke Here}, another participant named bowl \textit{Plate}, and another participant named stapler \textit{The Better Robot}. 

\subsection{Participants' Teaching Styles and Robot Performance}
\noindent We performed a three-way ANOVA with three independent variables: the two conditions (CBCL, FT), session number (total 5 sessions) treated as within subject, and previous robot programming experience of the participants (experts, and non-experts). The ANOVA was performed to understand the effect of the three independent variables on the teaching style of the participants and the robot's performance in the testing phase. The dependent variables were the classification accuracy of the robot in the testing phases for 200 sessions, the average number of images per object shown by the participants in each session, the number of teaching phases started by the participants in each session, and the number of times participants retaught misclassified objects in each session.  

Table~\ref{tab:anova_results} represents the $p$ values and significance levels for the ANOVA. For classification accuracy, we see a significant effect based on the session number, the choice of the CL model (CBCL or FT condition), and the interaction between the session number and the CL model. For the number of images taught per object, we noticed a significant effect based on the previous programming experience of the participants and the interaction between the CL model and the programming experience. For the number of teaching phases per session, we only saw a borderline effect by the programming experience and interaction between the CL model and the programming experience. Finally, reteaching of misclassified objects was not significantly affected by any independent variables. 
 
For significant ANOVAs, we performed the post hoc Tukey HSD test. However, the data for sub-groups for some dependent variables were not normally distributed, therefore, we also performed the Wilcoxon rank sum test \cite{Wilcoxon45Test} with false discovery rate correction \cite{benjamini1995controlling} for pairwise comparisons between sub-groups for each dependent variable.

\begin{figure}[t]
        \centering
    \begin{subfigure}{.5\linewidth}
        \includegraphics[width=\linewidth]{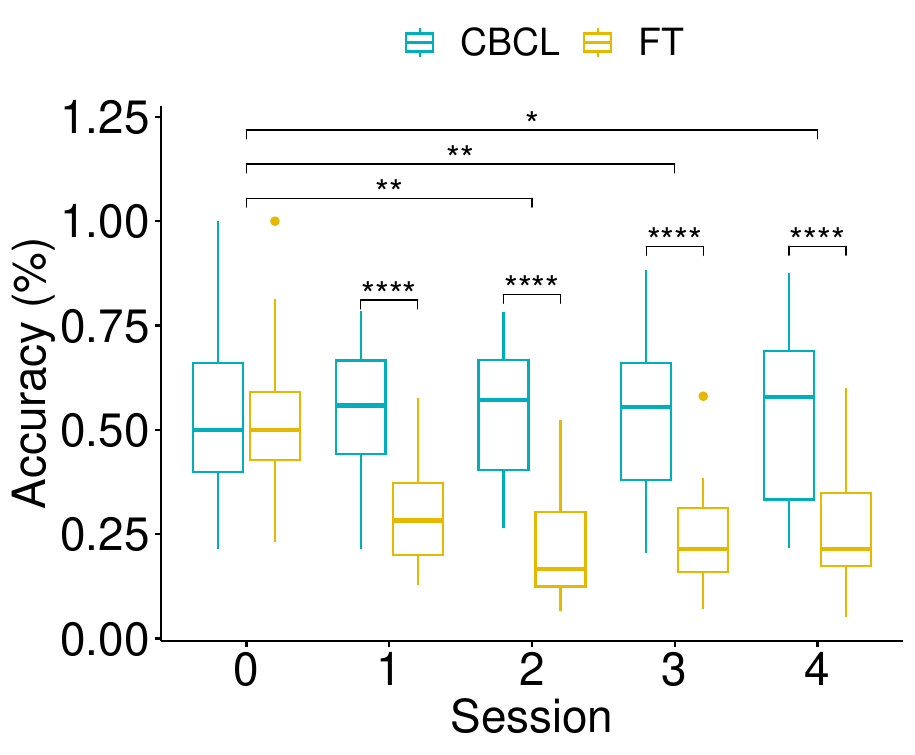}
        \caption{Accuracy of models by \\%
        session.}
        \label{fig:accu:overall}
    \end{subfigure}%
    \begin{subfigure}{.5\linewidth}
        \includegraphics[width=\linewidth]{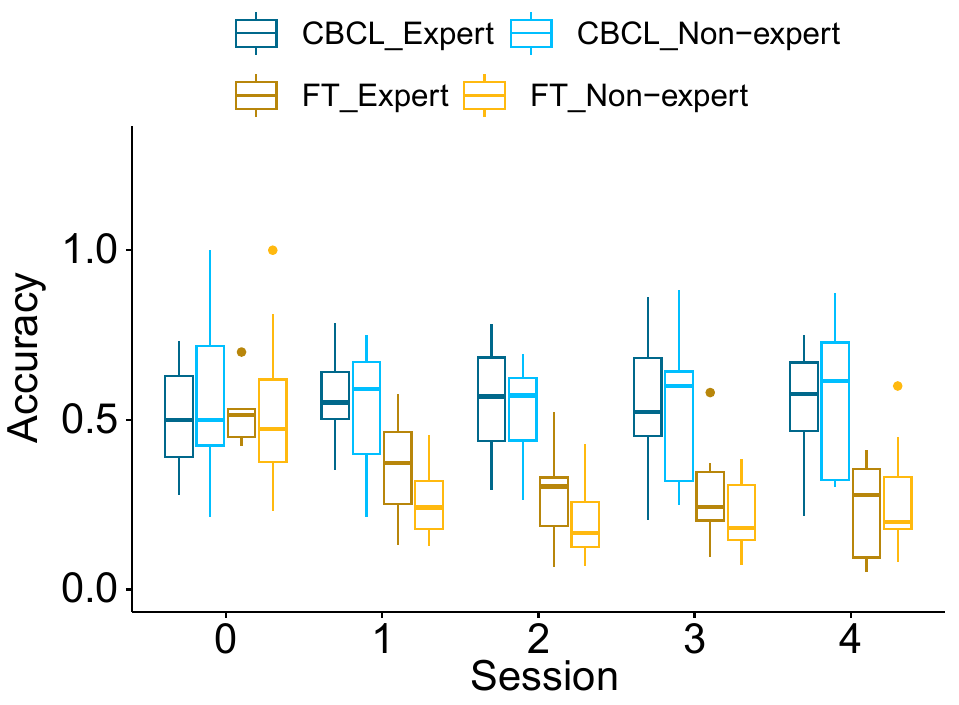}
        \caption{Accuracy of models for experts and non-experts.}
        \label{fig:accu:programming}
    \end{subfigure}
    \caption{Boxplots for \textit{accuracy} for two conditions. Significance levels ($* := p < .05; ** := p < 0.01; ****:=p<0.0001$) 
    are indicated on bars between columns.} 
    \label{fig:accu}
\end{figure}

\subsubsection{Model Accuracy}
As the data for classification accuracy was normally distributed, we performed the post hoc Tukey HSD test for significant ANOVAs. Figure \ref{fig:accu} shows the average classification accuracy of the continual learning robot over five sessions. As displayed in Figure \ref{fig:accu:overall}, the accuracy is significantly affected by the choice of the CL model. For the first session, both models have similar accuracy ($\mu=0.53$, $\sigma=0.19$ for CBCL; $\mu=0.52$, $\sigma=0.18$ for FT). For the next four sessions, there is a statistically significant difference between the two models: when comparing CBCL ($\mu=0.54$, $\sigma=0.16$) to FT ($\mu=0.29$, $\sigma=0.13$) with $p<0.0001$ for session 2, comparing CBCL ($\mu=0.54$, $\sigma=0.15$) to FT ($\mu=0.22$, $\sigma=0.13$) with $p<0.0001$ for session 3, comparing CBCL ($\mu=0.53$, $\sigma=0.20$) to FT ($\mu=0.24$, $\sigma=0.13$) with $p<0.0001$ for session 4, and comparing CBCL ($\mu=0.55$, $\sigma=0.19$) to FT ($\mu=0.26$, $\sigma=0.14$) with $p<0.0001$ for session 5. Further, when considering the two models separately, significant differences are seen between the first and the subsequent sessions for FT only. 

As evident from the ANOVA, there was no statistically significant difference in classification accuracy for expert and non-expert users (based on their previous programming experience). Results in Figure \ref{fig:accu:programming} correlate with the ANOVA. 

\begin{figure}
        \centering
    \begin{subfigure}{.5\linewidth}
        \includegraphics[width=\linewidth]{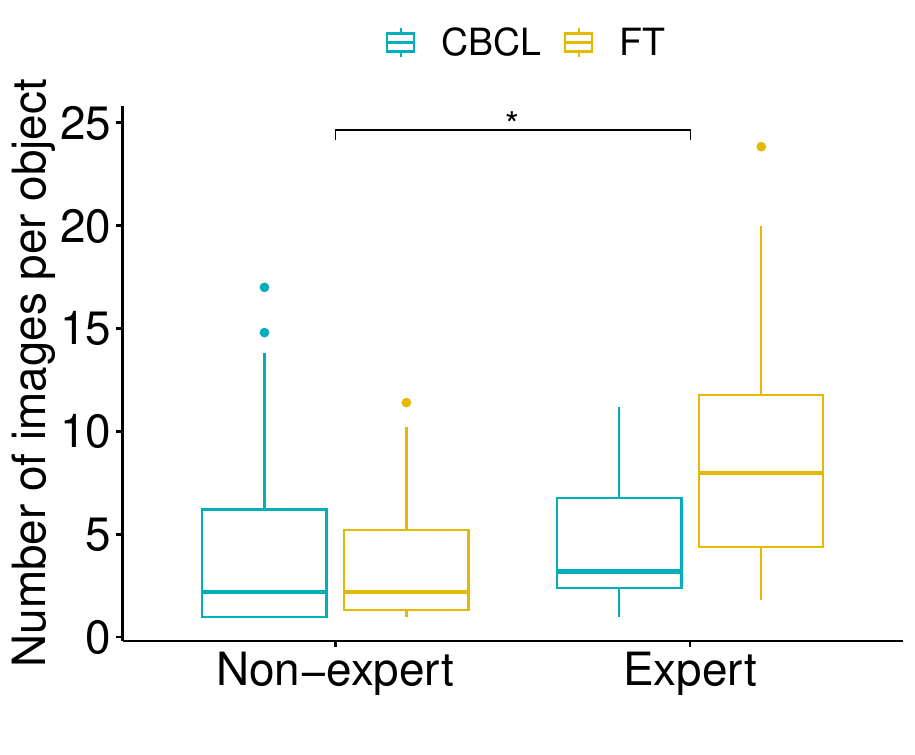}
        \caption{Number of images per object for experts and non-experts, \\
        and two conditions.}
        \label{fig:images:overall}
    \end{subfigure}%
    \begin{subfigure}{.5\linewidth}
        \includegraphics[width=\linewidth]{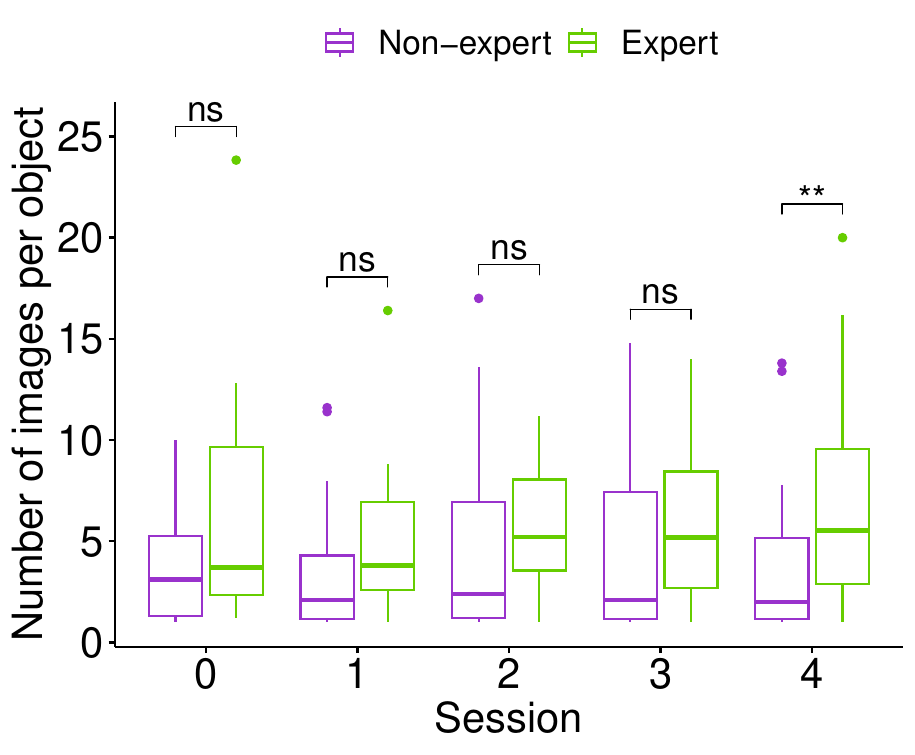}
        \caption{Number of images per object by session for experts and non-experts.}
        \label{fig:images:expert}
    \end{subfigure}
    \caption{Boxplots for \textit{number of images per object}. Significance levels ($* := p < .05; ** :=p < 0.01$) are indicated on bars between columns.} 
    \label{fig:images_overall}
\end{figure}

\subsubsection{Number of Images per Object}
We performed the post hoc Tukey HSD test for the significant ANOVAs for the number of images as the dependent variable. Figure \ref{fig:images:overall} details the difference between the two CL models and expert and non-expert participants in terms of the number of images taught per object. We notice a statistically significant difference between experts and non-experts irrespective of the CL model with ($\mu=3.93$, $\sigma=3.71$) for non-experts and ($\mu=6.09$, $\sigma=4.67$) for experts with $p=0.034$. However, this difference seems to stem from participants in the FT condition only. In terms of individual sessions, there is a statistically significant difference between experts (($\mu=6.75$, $\sigma=5.32$)) and non-experts (($\mu=3.91$, $\sigma=3.79$)) in session 5 only with $p=0.003$.  

\begin{figure}[t]
\centering
\includegraphics[width=0.7\linewidth]{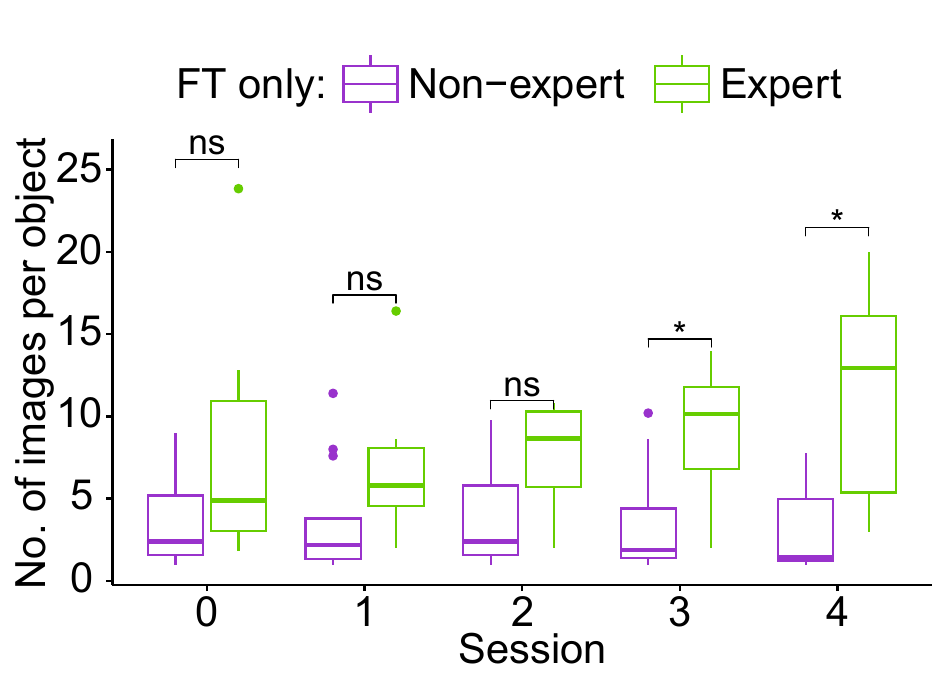}
\caption{\small Boxplot for \textit{number of images per object} for experts and non-experts in FT condition. Significance levels ($* := p < .05$) are indicated on bars between columns.} 
\label{fig:images_ft}
\end{figure}

To further investigate experts and non-experts in the FT condition, we performed a Wilcoxon rank sum test \cite{Wilcoxon45Test} between experts and non-experts in the FT condition over five sessions. As displayed in Figure \ref{fig:images_ft}, there is a statistically significant difference between experts and non-experts for sessions 4 and 5 only i.e. when comparing experts ($\mu=9.05$, $\sigma=4.41$) to non-experts ($\mu=3.68$, $\sigma=3.25$) with $p=0.035$, $W=10.5$ in session 4, and comparing experts ($\mu=11.49$, $\sigma=7.03$) to non-experts ($\mu=3.14$, $\sigma=2.49$) with $p=0.035$, $W=10.0$ in session 5. 

\subsubsection{Number of Teaching Phases per Session}
As the ANOVA for the number of teaching phases was not significant, we did not perform a post hoc Tukey HSD test. Overall, 20 out of 40 participants had at least one session where they started more than one teaching phase with the robot. Overall, 50 out of 200 sessions had more than one teaching phase ranging from 2 to 9 teaching phases in a single session.

\subsubsection{Reteaching after Misclassification}
The ANOVA for the dependent variable reteaching after misclassification was not significant and there were no borderline values. Therefore, we did not perform a post hoc Tukey HSD test. Overall, we noticed that 18 out of 40 participants retaught at least one object after it was misclassified by the robot during the testing phase. In total, there were 46 out of 200 sessions in which participants retaught misclassified objects with a maximum of 7 reteaching of misclassified objects. 

Note that the above statistic only counts the reteaching of misclassified objects from the current session only i.e. if an object taught in the previous sessions was misclassified and retaught in a session it is not covered in the above statistic. Overall, there were only 11 sessions when participants retaught at least one object from the previous sessions, with a maximum number of 4 old objects taught in a session. In terms of the number of participants, only 6 out of 40 participants retaught objects from previous sessions in subsequent sessions.

\section{Discussion}
\label{sec:discussion}
\noindent Results from the qualitative and quantitative analyses of the data collected in our study allow us to validate the hypotheses in Section \ref{sec:research_questions} and answer the research questions. 

For object labeling, we noticed significant variations in the labeling strategies of different participants. None of the 25 objects used in the study had a single consistent label across all 40 participants, even for simple objects, such as \textit{Apple}. Further, some participants also labeled different objects with the same label, and some participants gave multiple labels to the same object. These distinct labeling strategies seem to affect the performance of the continual learning robot as depicted by the high standard deviation in classification accuracy of both CL models (Figure \ref{fig:accu:overall}). As a consequence, we can accept \textbf{H1.1}. These results also indicate the need for developing personalized robots that adapt to their users' labeling strategies and learn, and understand, their environment such that both the user and the robot can effectively communicate about the entities in the environment. 

The classification accuracy of the continual learning robot was significantly affected by the choice of the CL model which was expected as the FT model forgets previous objects over the five sessions. However, classification accuracy was not affected by the previous robot programming experience of the participants. This result was surprising as it indicates that even expert users who have previous programming experience might not be familiar with continual learning over the long term. Therefore, the teaching effectiveness of both expert and non-expert users might be similar for a continual learning robot. Consequently, we have to reject \textbf{H4.1}. 

We quantified participants' teaching styles by calculating the number of images taught per object, 
the number of teaching phases started in each session, the number of times objects were retaught after being misclassified by the robot, and the number of times objects from previous sessions were taught by the participants. For the number of images per object, we did not find a statistically significant difference regarding the choice of the CL model or the session number in ANOVA. However, the previous robot programming experience of the participants did have a significant effect on the number of images per object. There was also an interaction of the previous robot programming experience with the session number and the CL model. Upon further investigation, we noticed that the difference between experts and non-experts occurred because expert users showed a significantly larger number of objects than non-expert users during later sessions in the FT condition. These results show that based on their previous experiences expert users might try to compensate for the degraded performance of the robot in later sessions by teaching more images per object. Note, that this might still not affect the robot's classification performance, as users might not be familiar with continual learning. 

In terms of the number of teaching phases per session, there was no statistically significant effect of the choice of the CL model or the session number. However, we did see that half of the participants started more than a single teaching phase in 200 sessions. Note that in the demo session participants were shown only a single teaching phase. Therefore, this result indicates that users might teach continual learning robots differently than the experimenter, i.e., not entirely following their instructions. 

For reteaching objects based on misclassification by the robot, we did not see any significant effect of the session number, choice of the CL model, or the previous programming experience of the participants. However, we did notice that almost half of the participants retaught objects if they were misclassified by the robot. This result indicates that, unlike static datasets, the continual learning robots might get more data for the objects if the robot misclassifies them in a session. Finally, we also noticed that almost half (45\%) of the participants also retaught some of the objects from previous sessions to the robot. Note that in the study instructions, and during the demo phase, participants were not told that they cannot re-teach old objects, therefore many of the participants retaught objects from previous sessions if the robot misclassified them during the testing phases. These results further demonstrate the difference between constrained CL test setups and testing in the real world with real users. Particularly, unlike constrained CL setups, users will reteach objects that they previously taught the robot if the robot does not classify them correctly. Finally, these results show that most users in the study were motivated to improve the performance of the robot, even though they were not given any specific incentive to do so. This is quite promising, as it indicates that users might be motivated to improve the performance of their personal robots over long-term interactions. Based on the above results \textbf{H2.1} can be accepted partially as we noticed that almost half of the participants retaught objects to the robot based on the robot's classification performance. \textbf{H2.2} (users teach a forgetful robot differently) has to be rejected as none of the three dependent variables for teaching style were affected by the choice of the model. Furthermore, \textbf{H3.1} (evolution of teaching styles over sessions) has to be partially accepted as we noticed a change in the number of images per object for expert users in the last session. Finally, we can accept \textbf{H4.2} (difference between the teaching styles of experts and non-experts) partially, as we did notice a difference in the number of images per object for expert and non-expert users. However, there was no difference in terms of the number of teaching phases per session and reteaching old objects between expert and non-expert users.  

\section{Conclusions}
\label{sec:conclusion}
\noindent In this paper, we considered a human-centered approach to continual learning to understand how users interact with and teach continual learning robots over the long term. We designed a long-term between-participant HRI study with a continual learning robot using two different CL models and analyze the data to understand the different teaching styles of participants, and how these styles are influenced by the performance of the robot over multiple sessions. Our results indicate that different users might teach household objects to the continual learning robot in a variety of ways, which could also affect the classification performance of the CL models. 
Moreover, the results show that the classification performance of the robot in prior sessions could influence the teaching style of the users in subsequent sessions, which is different from constrained CL test setups. The results also show that the previous programming experiences of the users can also significantly influence the way they interact with and teach the continual learning robot over multiple sessions. Finally, these results demonstrate the limitations of current CL test setups and CL models. Therefore, based on the results of this study, we recommend future CL models focus on adapting to the teaching style of their users, and that CL models should be tested in more realistic test setups. 

\section{Limitations and Future Work}
\label{sec:limitations}
\noindent We conducted our study in an unconstrained setup, where participants could teach and test the robot flexibly. However, the study was conducted in a robotics lab and not in a realistic household environment. In future work, we plan to conduct a similar study in a smart home with the same robot to understand the influence of the household environment on the interactions and teaching styles of the users. 
We conducted the user study with a mix of expert and non-expert users, however, they were all university students between the ages of 18 and 37 years. In future work, we plan to conduct this study with participants who might be less familiar with robots to understand the effectiveness of continual learning robots for assistive applications. Finally, the study was conducted with one particular robot and with two CL models. Expanding this work to other robots and CL models can help us understand the larger design space of continual learning robots and users' teaching patterns when interacting with these robots.

Despite these limitations, our user study took the first step toward a human-centered approach to continual learning by integrating machine learning-based CL models with HRI. We hope that our results can help ML and HRI researchers design CL models that can adapt to their users' teaching styles and test these models in realistic experimental setups where embodied agents interact with human users.

{\small
\bibliographystyle{IEEEtran.bst}
\bibliography{main}
}

\end{document}